# SikuGPT: A Generative Pre-trained Model for Intelligent Information Processing of Ancient Texts from the Perspective of Digital Humanities

Chang Liu[1], Dongbo Wang[1], Zhixiao Zhao[1], Die Hu[1], Mengcheng Wu[1], Litao Lin[1], Si Shen[2], Bin Li[3], Jiangfeng Liu[1], Hai Zhang[1], Lianzheng Zhao[4]

1 College of Information Management, Nanjing Agricultural University, Nanjing 210095, China
2 Group of Science and Technology Full-text Knowledge Mining, School of Economics & Management, Nanjing University of Science and Technology, Nanjing 210094, China
3 College of Liberal Art, Nanjing Normal University, Nanjing 210097, China
4 School of Foreign Languages, China Pharmaceutical University, Nanjing 211198, China

**Abstract:** The rapid advance in artificial intelligence technology has facilitated the prosperity of digital humanities research. Against such backdrop, research methods need to be transformed in the intelligent processing of ancient texts, which is a crucial component of digital humanities research, so as to adapt to new development trends in the wave of AIGC. In this study, we propose a GPT model called SikuGPT based on the corpus of *Siku Quanshu*. The model's performance in tasks such as intralingual translation and text classification exceeds that of other GPT-type models aimed at processing ancient texts. SikuGPT's ability to process traditional Chinese ancient texts can help promote the organization of ancient information and knowledge services, as well as the international dissemination of Chinese ancient culture.

## 1 Introduction

In 2022, generative AI has made remarkable achievements in various fields. The diffusion models that automatically generate images, cross-modal models that generate videos with one click, and the astounding ChatGPT all showcase the charm of AIGC (AI-Generated Content). The China Academy of Information and Communications Technology pointed out in its "White Paper on Artificial Intelligence-Generated Content (AIGC) "(*White Paper on Artificial Intelligence-Generated Content (AIGC) (2022) -- China Information and Communication Academy*, n.d.) that AIGC will become a unique information production method in the web3.0 era, following PGC (Professionally Generated Content) and UGC (User Generated Content). Large-scale pre-trained models have officially entered people's lives as productivity tools, bringing about tremendous changes to the industry.

The meeting of ancient texts and generative AI has injected new vitality into the intelligent processing of ancient texts. Using pre-trained models to generate texts can assist in the interlingual translation and intralingual translation of ancient texts and knowledge organization work. Currently, large language models represented by ChatGPT have been able to achieve barrier-free interaction with humans and can complete different natural language processing tasks according to human instructions. However, they may yield higher error rates when answering certain vertical field questions, because the training process of large language models focuses more on learning general knowledge than specialized knowledge. At the same time, using subject knowledge for fine-tuning or writing prompts for context learning is too costly, and domain-specific deployment in real scenarios remains very challenging. Previous research (Moradi et al., 2022) has indicated that in text processing in the field of biology, it is difficult for GPT3 to exceed the ability of small models in solving small sample tasks under the condition of the same corpus pre-training and fine-tuning. Using domain data to pre-train a language model with billions of parameter can significantly enhance the data mining and generation capabilities in low-resource environments.

In this study, we developed a SikuGPT pre-trained model based on the GPT2 model for the automatic generation of ancient texts. We verified the performance of the generative model in two categories of tasks: text translation and text comprehension. In this article, we also released an open-source Ancient Chinese-Modern Chinese translation model fine-tuned with a bilingual parallel corpus, with the twofold aim of promoting research efficiency for scholars working in Chinese ancient texts and boosting the international dissemination of Chinese ancient culture.

## 2 Related Research

### 2.1 Generative Pre-trained Language Models

After going through statistical learning models and deep neural network models, NLP technology has officially entered the era of pre-trained language models. Language models with the Transformer (Vaswani et al., 2017) structure as the basic framework promote the reasonable allocation of information and computing resources by training a general text representation using a large-scale unlabeled corpus, and then using it for the intelligent processing of downstream text with similar language features. The "pre-training + fine-tuning" or "pre-training + template prompt + fine-tuning" information processing process has officially become the basic paradigm of NLP research in the new era (Liu et al., 2023). The existing diverse pre-trained models can be roughly divided into autoencoder models, autoregressive models, and sequence-to-sequence models according to their basic architectures. Generally speaking, Autoencoder models only stack Transformer encoder structures, construct loss functions by predicting masked tokens in a sentence and excel at text comprehension tasks. Relevant models include BERT (Devlin et al., 2019), ERNIE (Zhang et al., 2019), and etc.

Autoregressive models and sequence-to-sequence models are mainly text generation models. Autoregressive models merely stack Transformer decoder structures, thus being suitable for unidirectional text continuation. During training, the model needs to predict the probability distribution of the output vocabulary in the vocabulary table based on the input partial vocabulary information. In addition, it has to calculate the loss function by comparing the predicted results with the content of the original sequence, and then complete parameter updating. OpenAI, which was the first to use Transformer decoders for text feature extraction, released the GPT1 model (Radford, Narasimhan, et al., n.d.) in 2018. With excellent generation ability, this model can also be applied to natural language understanding tasks by changing the task paradigm. GPT2 (Radford, Wu, et al., n.d.) inherited the basic structure of GPT1, but improved the input form of training data, making the style of pre-training data and task data more similar. GPT2 utilizes more parameters and larger training texts, and the largest-scale GPT2 model has 1.5 billion parameters, which grant it capabilities of zero shot learning. ChatGPT, which has attracted wide attention from scholars recently, represents an improved version of GPT3 (Brown et al., 2020) as the language infrastructure. GPT3 has 96 Transformer decoders, each containing 180 million trainable parameters. Containing 175 billion parameters, the GPT3 model uses 45TB of data during training. Compared with small models, large-scale language models have the phenomenon of "emergence". In systems science, "emergence"(Holland, 2000) is used to describe the phenomenon or property that the interaction between subsystems within a system produces features that individuals do not possess. This phenomenon of "emergence" in language model training (Wei et al., 2022) is manifested as the fact that language models larger than a certain threshold have intelligence that ordinary models do not have. As one of the largest language models, GPT3 can model all NLP tasks generatively. Even in the case of providing small samples or no sample at all, the GPT3's performance in answering questions is close to or exceeds the upper limit of the fine-tuning ability of small-to-medium-sized models. OpenAI has successively launched InstructGPT (Ouyang et al., 2022) and ChatGPT based on the iterative model GPT3.5, which comprehensively applies technologies such as code training, instruction-tuning, and human feedback reinforcement learning to further stimulating the potential of large-scale language models. This ultimately enables the model to generate high-quality answers correctly according to human instructions. In order to break the monopoly of large model services, Huggingface led the BigScience project, which organized over 1,000 scientists from different backgrounds around the world to jointly train and open-source the BLOOM model (Workshop et al., 2023). This model has a structure and scale similar to GPT-3, but lower pre-training data volume; despite this, its multi-disciplinary background and multi-task learning still give it good small-sample learning ability.

As another major type of generative model, sequence-to-sequence models share the Transformer structure as a whole or use separate encoder and decoder structures to extract text features. They excel at performing sequence-to-sequence tasks such as automatic summarization, text translation, and automatic question answering where the length of input and output sentences may vary. With the inclusion of an encoder structure, the language understanding ability of sequence-to-sequence models is generally better than that of auto-regressive models within a certain range. Representative achievements in this type of model include the MASS (Song et al., 2019) and Unilm (L. Dong et al., 2019) models released by Microsoft, the BART (Lewis et al., 2019) model proposed by Facebook, and the T5 (Raffel et al., 2020) model proposed by Google. The MASS model, combining the advantages of BERT and GPT, possesses a basic structure of "encoder-attention-decoder". During pre-training, MASS masks k consecutive words on the input side and adopts the decoder to predict the masked vocabulary, thus avoiding the problem of inconsistent parameters caused by separately optimizing the encoder and decoder. Unilm directly reuses the bidirectional encoder structure of the original BERT model and exploits three mask matrices to unify the pre-training tasks of bidirectional language models, unidirectional language models, and sequence-to-sequence models. The BART model also absorbs the advantages of BERT and GPT. As a result, it owns the following feature: after allowing different noises to be used to corrupt the text, it adopts an "encoder-decoder" structure to reconstruct the text, rather than just applying masked vocabulary to pre-training, which helps the model better understand the information of the sequence itself. The T5 model, one of the hottest language models in recent years, own as its basic structure a Transformer that has improved layer normalization and position encoding. T5's pre-training process is divided into two stages: self-supervised learning and supervised learning. In the self-supervised learning stage, T5 differs from models like BERT that uses the MLM pre-training task. The reason is that each masked target represents a continuous text interval rather than a single vocabulary. Further, the target that the decoder needs to restore is only the masked content rather than the sentence itself. The above steps are grouped into input text and output results by assigning a sentry token. In the supervised learning stage, the researchers of T5 transformed all conventional NLP tasks into "text-to-text" generation tasks and then added a specific prefix to the input data of each task. This gave the model the ability to generate specific texts based on human instructions and made the data format of contextual tasks more suitable for parameter optimization. Similarly, sequence-to-sequence models are also commonly selected architectures in large model production. Related studies such as the FLAN-T5 (Chung et al., 2022) and T0 (Sanh et al., 2022) models have re-optimized the pre-training of the T5 model. One of their features includes using instruction tuning to achieve contextual learning with small sample data.

## 2.2 Applications of Generative Pre-trained Language Models in Intelligent Processing of Ancient Texts

Generative pre-trained models can be combined with ancient text corpora or multilingual parallel corpora to perform automated translation, text summarization, automatic question answering, text completion, and other generation tasks. They can also be used to perform natural language understanding tasks such as text classification, information extraction, and text retrieval by restructuring the input and output patterns of these tasks. The following sections will introduce the relevant research on the applications of generative pre-trained models in processing ancient texts.

### 2.2.1 Generative Models and Ancient Text Generation

Generating ancient texts is one of the core applications of generative models. Ancient text generation tasks can be classified into single-language generation and cross-language generation depending on the generation target. Single-language generation requires that the generated text is in the same category as the original text, while cross-language text generation is the opposite. In fact, due to the significant differences between the ancient and modern forms of many languages, intralingual text translation can be viewed as a special case of cross-language text generation.

Examples of single-language generation tasks include generating poetry and ancient language. Relevant studies include Liao et al. (2019) who used GPT models to generate Chinese classical poetry in different styles. Hu & Sun (2020) built a unified framework for generating Chinese classical poetry based on the GPT2 model, using a form of weighted emphasis to control the style of the generated text. The relevant results were included in the "Nine Songs" (Zhipeng et al., 2019) poetry generation system developed by Tsinghua University. Nguyen et al. (2021) developed a Vietnamese poetry generation model called SPGPT2 based on GPT2, which generates Vietnamese traditional poems that conform to the Luc Bat format by adding constraints.

Cross-language text generation tasks include ancient text translation and text summarization. Yang et al. (2021) and Tian et al. (2021) exploited the UNILM framework to load GUWEN-BERT (Ethan, 2020/2023) and AnchiBERT (*Xujiacheng127/Anchi-Bert · Hugging Face*, n.d.) models pre-trained on simplified ancient Chinese data for cross-language text generation tasks. Comparative experimental results showed that adding pre-training mechanisms can effectively improve the model's generation ability. Chang et al. (2021) designed a translation interface that is sensitive to temporal information based on the GPT2 model to address translation quality problems caused by temporal differences in ancient texts. Jin et al. (2022) constructed a bilingual parallel corpus of Chinese classical poetry and modern Chinese translations. By use of neural machine translation and trained Transformer and GPT2 models based on the translation data, they achieved automated translation of Chinese classical poetry. Peng et al. (2021) constructed a historical text dataset in German and Chinese and proposed an algorithm for summarizing ancient historical texts into modern written language using cross-language transfer technology.

**2.2.2 Generative models and text completion**

Owning to their pre-training tasks, generative pre-training models, including language models and multimodal models, can repair noisy targets and complete missing information. Some researchers utilize generative models to restore the ancient texts which suffered physical damages for various reasons. For example, Fetaya et al. (2020) trained a recursive neural network with digitized cuneiform corpora from Mesopotamia to repair damaged Babylonian texts. Assael et al. (2022) used Transformer as the basic structure to develop a model to restore damaged ancient Greek inscriptions. The model accepted text at the level of single characters and designed other structures to model the temporal and geographical characteristics of the text, achieving restoration results better than human experts. Wenjun et al. (2023) developed a double-branch character restoration network based on generative adversarial networks. They exploited image data to train two branches to extract the basic features of the damaged characters and example characters, achieving good restoration results.

In addition to generating ancient texts, generative pre-training models may even change the basic paradigm of natural language understanding tasks. The current label classification problem may gradually shift to specific character generation problems. In theory, a large enough generative model can be applied to all natural language processing tasks, but this requires the joint efforts of multidisciplinary researchers to share discipline data and task descriptions. However, the use of Autoencoder models, represented by BERT, for knowledge extraction in ancient texts remains the mainstream in datamining studies of ancient books. More attention deserves to be paid to the application of generative models in research related to the mining of ancient texts.

# 3 Research Methods

## 3.1 Pre-training Data Source

In this study, the data used for pre-training is the *Siku Quanshu* of Wenyuan Ge version, which is a large series of books compiled during the reign of Emperor Qianlong in the Qing Dynasty, including four parts: "Jing" Part (经部, classical literature), "Shi" part (史部, historical literature), "Zi" part (子部, ideological literature), and "Ji" part (集部, literary works). Previously, our research team constructed the SikuBERT and SikuRoBERTa models based on the Wenyuan Ge's *Siku Quanshu*, which showed superior performance in tasks such as ancient Chinese part-of-speech tagging, text segmentation, and named entity recognition (Wang et al., 2022). As the largest collection of ancient texts in China, the *Siku Quanshu* covers a wide range of content and spans a long period of history. It is not only a vital resource for studying ancient Chinese classics, but also an important reference for understanding Chinese culture and history. In this study, we choose the traditional Chinese format of the *Siku Quanshu* texts as the training corpus to better align with the original texts of ancient books.

## 3.2 Pre-training Method

### 3.2.1 Pre-training Model Selection

The pre-training models selected in this experiment are as follows: (1) GPT2-Chinese-ancient model (*Uer/Gpt2-Chinese-Ancient · Hugging Face*, n.d.) open-sourced by Huggingface, which was trained on ancient Chinese literature from the Daizhige library, with a total of 3,000,000 Chinese characters. The training process is based on the UER (Zhao et al., 2019) open-source framework, and the model consists of 12 layers with a vocabulary size of 25,370. (2) GPT2-base-Chinese (*Ckiplab/Gpt2-Base-Chinese · Hugging Face*, n.d.) model introduced by the CKIP Lab, which is a GPT2-like model designed for traditional Chinese.

### 3.2.2 Pre-training method selection

As a typical autoregressive language model, GPT has a unidirectional Transformer decoder structure and adopts the training method of the causal language model (CLM). The causal language model is a pre-training task for training unidirectional text representations, in which the model only needs to predict the next word based on the vocabulary on one side of the input sentence and then use a cross-entropy loss function to update the model parameters. Compared with masked language model, the causal language model only allows reference to one side of the content for prediction. When the training goal is to learn a good representation of the input text, masked language model (MLM) is undoubtedly a better choice due to its ability to consider context simultaneously. However, when the training goal is to generate fluent text, the unidirectional causal language model is similar to human writing and can better improve the model's creativity. In this experiment, the training task adopts the causal language model (CLM) and is completed by using the Transformers framework provided by Huggingface company.

## 3.3 Downstream task design

To verify the performance of the SikuGPT pre-training model, this study selects two GPT models designed for ancient Chinese text processing, GPT2-Chinese-ancient and GPT2-base-Chinese. They are chosen as baseline models to further verify their performance in two natural language processing tasks: ancient text translation and ancient text classification.

### 3.3.1 Text Translation

Machine translation is an important task in the field of natural language processing. The language model's ability to understand language can indirectly reflect the model's machine translation performance. At the same time, machine translation can test the model's generation ability. There is still much room for improvement in machine translation for ancient texts because of their unique grammar and vocabulary, especially traditional Chinese texts. As an essential part of Chinese classic literature, standard language and outstanding literary attainments make Twenty-Four Histories an excellent corpus for machine learning testing. This study selects the ancient Chinese text alignment corpus of Twenty-Four Histories as the evaluation corpus. BLEU is adopted as the evaluation metric to test the model's understanding and generation ability.

### 3.3.2 Text Classification

As a basic task in the field of natural language processing, text classification aims to classify a given text into predefined categories. Unlike machine translation tasks, text classification tasks are simpler and more direct. On the one hand, text classification tasks usually require understanding the semantics of the text. Text classification tasks can be applied to testing the model's ability to understand natural language meanings. On the other hand, the model proposed in this study is a language model obtained through unsupervised training.

*Siku Quanshu* includes four major categories, each of which is further divided into multiple sub-categories. In this study, the sub-categories in Classics are selected as the test corpus, with the classification labels based on the content of the sub-categories. Precision, recall, and F-score are utilized as evaluation metrics to test the model's understanding and generalization ability.

## 4 Experimental Procedure and Results

### 4.1 Pre-training Experiment

All the Chinese characters that can be displayed in utf-8 encoding were extracted from the entire text of the *Siku Quanshu*, with 5086 characters added to the GPT2-chinese-ancient model's vocabulary. The filtered *Siku Quanshu* text was divided into a training set and a validation set in a ratio of 99:1. The model was then fine-tuned using the CLM method based on the Transformers framework. The model training parameters are shown in Table 1.

**Table 1 Key hyperparameters for model training**

| Hyperparameters | Value |
|---|---|
| learning_rate | 5e-5 |
| num_train_epochs | 3 |
| per_device_train_bach_size | 8 |

For preliminary evaluation of the model's performance, perplexity was chosen as the evaluation metric. Perplexity is a language performance evaluation metric based on the probability of sentences in the validation set. The rationale is that a language model performs better if it assigns higher probability values to sentences in the validation set. Since the sentences in the validation set are all normal sentences, a well-trained language model that assigns higher probabilities to the sentences in the validation set indicates a better fitting ability of the model. In perplexity calculation, a sentence can be represented as:

$$S = W_1, W_2, ..., W_n \#(1)$$

The appearance probability of a sentence is:

$$P(S) = P(W_1, W_2, ..., W_n) = P(W_1)P(W_2|W_1)...P(W_n|W_1, W_2, ..., W_{n-1})\#(2)$$

The formula for calculating perplexity is based on the formula for calculating the probability of a sentence:

$$PPL = P(w_1w_2...w_n)^{-\frac{1}{n}} = \sqrt[n]{\frac{1}{P(w_1w_2...w_n)}}\#(3)$$

We used three GPUs of type RTX 8000 to complete the pre-training task of the model, the whole process took 3 days. The perplexity score of the pre-trained SikuGPT model is 20.85, which does not appear to be a good indicator of performance. However, compared to Autoencoder models, Autoregressive models typically have higher perplexity scores. Moreover, perplexity is not the only standard for evaluating model performance, but rather merely one of the initial indicators, and the performance of the model needs to be explained through its performance in downstream tasks.

## 4.2 Downstream Task Verification

### 4.2.1 Text Translation Task

Based on the CLM task, the pre-training language model SikuGPT can predict the probability distribution of the next word or sequence given the previous context information. Similar to language modeling, the goal of the CLM task is to learn contextual and semantic information of language, so the pre-trained model trained on the CLM task usually has good language understanding and generation abilities. To test the practical effectiveness of the SikuGPT pre-training model, this study used the twenty-four histories parallel corpus of ancient and modern Chinese as experimental data. Afterwards, the machine translation task was exploited to evaluate the model's actual performance. Experimental controls included the GPT2-chinese-ancient and GPT2-base-chinese pre-trained models, as well as the traditional single-layer Transformer model.

**(1) Data and Task Description**

The data in this study consists of ancient Chinese sentences and their corresponding modern Chinese translations from the Twenty-Four Histories. The Twenty-Four Histories is the general term for the twenty-four official Chinese historical records compiled by scholars in various dynasties in ancient China. It covers nearly 5,000 years of history from the Yellow Emperor period to the Ming Dynasty, and contains a diversity of ancient Chinese culture, including politics, economy, military, and thoughts and etc., making it a valuable cultural heritage of human civilization. The aligned corpus in the text translation task in this study comes from

China's "The 11th Five Year Plan" key work, "The Complete Translation of the Twenty-Four Histories(《二十四史全译》)". This was jointly compiled by more than 200 experts in ancient books research over a period of 13 years, representing the highest quality of modern Chinese translations of the Twenty-Four Histories. OCR (Optical Character Recognition) technology and the Aligner alignment software was utilized to align the ancient Chinese and modern Chinese texts at the sentence and paragraph levels. Subsequently, the aligned sentences were filtered based on named entity recognition technology and text matching technology to ensure accurate entity annotation in both ancient and modern Chinese sentences. The aligned sentences with a similarity score between ancient and modern Chinese sentences of 0.85 to 0.98 were selected according to similarity detection. In total, 300,000 high-quality aligned sentence pairs were selected from nearly 1 million aligned sentence pairs for experimentation.

The character lengths of the ancient and modern Chinese sentences total 9,368,674, and 12,236,739, respectively. Their average sentence lengths reach 30.47 characters, and 39.80 characters, respectively, indicating that ancient Chinese is more concise in description than modern Chinese. In terms of variance, ancient Chinese sentences is 1977.29, while modern Chinese sentences is 3424.18, indicating that there is greater dispersion between sentences in modern Chinese compared to ancient Chinese.

These details are shown in Table 2. The average sentence length of ancient Chinese in the corpus is over 30 characters, which is to enable the model to better learn the relationship between parallel sentence pairs in the corpus. In addition, during the preprocessing of the experimental data, the corpus was divided into a training set and a test set in a 9:1 ratio. Specifically, 270,000 pairs were used as training data, while 30,000 pairs were used as test data. A sample of the aligned ancient-modern Chinese sentence pairs is shown in the Table 2.

**Table 2 Basic information of ancient-modern Chinese alignment corpus**

|  | Ancient Chinese | Modern Chinese |
|---|---|---|
| Total number of characters | 9368674 | 12236739 |
| Average word count per sentence | 30.47 | 39.80 |
| variance | 1977.29 | 3424.18 |

**Table 3 Examples of aligned corpus**

| Corpus type | Corpus type |
|---|---|
| ancient-modern Chinese alignment corpus | {"Ancient": "後與秦戰，爲秦所獲，立十四年而死。", "Chinese": "後來與秦國作戰，被秦軍捉住，在位十四年而死。"} |

*"後來與秦國作戰，被秦軍捉住，在位十四年而死。"means " Later, he fought against the state of Qin and was captured by the Qin army. He reigned for 14 years and died."

**(2) Model Validation**

To validate the performance of SikuGPT, the baseline models selected for the validation experiment are GPT2-chinese-ancient, GPT2-base-chinese, SikuBERT, and Transformer

GPT2-chinese-ancient is a pre-trained language model based on GPT2 for generating ancient Chinese text. It is trained on un-punctuated data from the DaiZhige ancient Chinese corpus and then supplemented with punctuation. GPT2-base-chinese, a Chinese version of GPT2 trained on 15GB of Chinese language data, can be exploited for poem generation, news generation, and novel continuation. SikuBERT is a pre-trained model based on the BERT-base-Chinese architecture and fine-tuned on the *Siku Quanshu* corpus by Nanjing Agricultural University. With a word list using traditional Chinese without punctuation, sentence segmentation is performed at the character level. To adapt the BERT-based model for text translation tasks, the unilm framework is adopted to load the model weights and reconstruct the BERT model's input format.

**(3) Model Validation Performance Evaluation Metrics**

BLEU (Bilingual Evaluation Understudy) is a commonly used evaluation metric in machine translation. BLEU measures the similarity between the machine translation output and the human reference translation to evaluate the quality of machine translation systems. The evaluation result is represented as a score between 0 and 1, with higher scores indicating better machine translation quality. BLEU metric calculation is based on n-gram matching and sentence length penalty. Specifically, the BLEU metric matches the n-grams in the translation output with those in the reference translation and calculates an n-gram matching score based on the number of matches. Meanwhile, the BLEU metric penalizes longer translation results to avoid machine translation systems generating meaningless vocabulary to increase their scores.

**(4) Model Hyperparameter Settings**

The computer configuration in this experiment is as follows: Operating System: CentOS 3.10.0; CPU: 4 Intel(R) Xeon(R) CPU E5-2650 v4 @ 2.20GHz; Memory: 256G; GPU: 6 NVIDIA Tesla P40, VRAM: 24G. To ensure the comparability of the validation results, the hyperparameters of the model training are kept consistent in the validation experiment. The key hyperparameters used in text translation tasks are shown in the following table:

**Table 4 Key hyperparameters of translation model training**

| Hyperparameters | Value |
|---|---|
| max_gen_length | 512 |
| max_seq_length | 1024 |
| train_batch_size | 8 |
| epoches | 5 |
| leraning_rate | 1e-5 |
| warmup_proportion | 0.1 |

**(5) Comparison of Text Translation Performance**

As shown in Table 5, in terms of Chinese text translation performance, the GPT-based models all surpass the basic Transformer architecture. Both SikuGPT and GPT2-base-chinese models perform better than GPT2-chinese-ancient and SikuBERT+unim models in terms of single-character level and fluency of generated text. Furthermore, the translation performance of the GPT-based models exceeds that of SikuBERT in terms of both single-character level and fluency. Overall, SikuGPT performs the best in the translation task with slight advantages in every BLEU score compared to GPT2-base-chinesegpt2chinese. In the translation task, SikuGPT not only captures the semantic information of words and phrases accurately but also performs better in capturing the semantic information of longer sentences, resulting in more accurate translations.

**Table 5 Comparison of translation performance of different models**

| Model | BLEU-1 | BLEU-2 | BLEU-3 | BLEU-4 |
| --- | --- | --- | --- | --- |
| GPT2-chinese-ancient | 0.735 | 0.546 | 0.432 | 0.351 |
| SikuGPT | 0.765 | 0.593 | 0.488 | 0.413 |
| GPT2-base-chinese | 0.761 | 0.589 | 0.485 | 0.410 |
| SikuBERT+unilm | 0.658 | 0.514 | 0.423 | 0.357 |
| Transformer | 0.695 | 0.500 | 0.379 | 0.301 |

To evaluate the specific translation performance of each model, we randomly selected a test sentence from the test set, such as "後與秦戰，爲秦所獲，立十四年而死." The specific translation results are shown in Table 6. Overall, the translation results of the randomly selected test sentence basically retain the semantic information of the original text. However, regarding special grammatical phenomena such as polysemy and proper nouns, the translation performance is not satisfactory. For instance, "秦" can refer to both the state of Qin and the Qin army, which increases the difficulty for model learning. Feasible solutions may include enlarging the scale of training corpus, further integrating text part-of-speech features, or fusing more knowledge in the field of classical Chinese. It is worth mentioning that where there is no given examples, the translation accuracy of ChatGPT for this sentence is not ideal. Thus, substantial data still needs to be consumed for the model's context learning if ChatGPT is to be applied to classical Chinese text translation.

**Table 6 Translation results**

| Model | Translation results |
| --- | --- |
| GPT2-chinese-ancient | 後來和秦軍作戰，被秦軍俘獲，立下十四年死去。 |
| SikuGPT | 後來與秦軍交戰，被秦國擒獲，在位十四年死去。 |
| GPT2-base-chinese | 後來和秦作戰，被秦俘獲，在位十四年而死。 |
| SikuBERT+unim | 後來與秦軍交戰，被秦俘獲，在位十四年後死去。 |
| ChatGPT(zero-shot) | 後與秦國交戰，被秦國俘虜，後來在秦國度過了十四年並最終去世。 |
| Transformer | 後來與秦國交戰 ，被秦國獲得，立十四年而死 。 |

### 4.2.2 Text Classification Task

To further validate the performance of SikuGPT pre-training model, a text classification task was also conducted in this study as a downstream validation experiment. In this experiment, the required training and testing data was constructed based on the 14 categories of ancient texts in the sub-parts of *Siku Quanshu*. Several pre-trained models were introduced as baseline models for the validation experiment, including GPT2-chinese-ancient and GPT2-base-chinese as well as SikuRoberta, which was specifically fine-tuned for the ancient text classification task.

**(1) Corpus and Task Description**

*Siku Quanshu* classified ancient Chinese books into four categories, and the "Zi" part were further divided into 14 sub-categories, including Confucianism, Military Strategists, Legalism, Agriculturalism, Medicine, Astronomy and Mathematics, Arts, Divination and Numerology, Rituals and Records, Miscellaneous Works, Novels, Buddhism, Taoism, and Genus-books. The corpus used in the text classification task was the traditional Chinese version of *Siku Quanshu* data collected through web crawlers (Hu et al., 2022). After preliminary data preprocessing, a total of 132,315 valid text data containing all 14 categories were obtained. Afterwards, the data was divided into training and testing sets in a ratio of 9:1, with 119,083 as training data and 13,232 as testing data. The distribution of training and testing data in each sub-category is shown in the following figure.

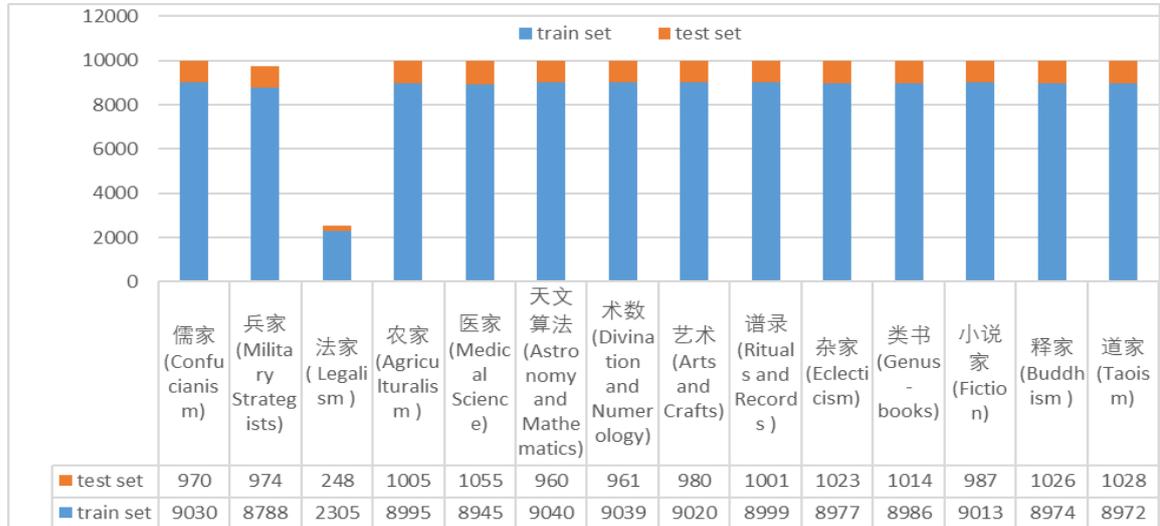

**Figure 1 Distribution of text classification task data set across sub-categories**

The text classification task aims to make the GPT pre-training model automatically generate the category labels correctly through training fine-tuning on the above corpus. Generally, the idea of utilizing pre-trained models for text classification is to input the output vector of the model's last layer into the fully connected layer to calculate the probability of each category. For GPT pre-training models, the output vector of the model is the vector corresponding to the last token of the input sequence. From the training task, this vector represents the semantic information of the token which is most likely to correspond to the next character under the condition that the first n tokens are determined. If directly exploited for text classification, this vector may cause a problem of inconsistency between the downstream task and the pre-training task structure. Therefore, the research team borrowed the idea of prompt and used the generative modeling approach to restructure the task paradigm of text classification. Specifically, a concatenation prompt sentence was added at the end of the input text to guide the model to decode the text of the label itself, thus converting a classification task into a generation task, and ensuring the consistency of the upstream and downstream training tasks. The template design method is shown in

**Table 7 Prompt template design**

| Original statement | Prompt template | Input statement |
|---|---|---|
| [CLS]text[SEP] | 这个句子的类别是____ | [CLS]text[SEP] 这个句子的类别是____ |

*"这个句子的类别是____"means "The category of this sentence is____".

**(2) Model performance evaluation index**

For the performance evaluation of each category in text classification, precision (P), recall (R), and F1-score are adopted as evaluation metrics. Weighted precision (Weighted_P), weighted recall (Weighted_R), and weighted F1-score (Weighted_F) are used to calculate the overall classification performance.

**(3) Comparison of text classification effect**

According to Table 8, in terms of overall classification performance, SikuGPT performed the best among the four models with all metrics reaching over 90%, and achieved a slight advantage over the 1.0 version of the SikuRoberta model trained with fine-tuning. GPT2-chinese-ancient performed poorly, with a large gap in all metrics compared to the other three models. This may be attributed to the fact that GPT2-chinese-ancient training data is in simplified Chinese characters, thus unable to handle traditional Chinese characters well. GPT2-base-chinese performed slightly worse than SikuGPT. Even though it was trained on modern Chinese language data, the model still performed well in classification. This also demonstrates the importance of maintaining consistency between simplified and traditional Chinese language data in building ancient language models.

**Table 8 Overall classification effect comparison of each model**

| Model | Precision（%） | Recall（%） | F1-score（%） |
|---|---|---|---|
| GPT2-chinese-ancient | 88.34 | 88.38 | 88.29 |
| GPT2-base-chinese | 90.25 | 90.24 | 90.23 |
| SikuGPT | 90.34 | 90.35 | 90.30 |
| SikuRoberta (non-prompt) | 86.37 | 94.60 | 90.21 |

The detailed precision, recall, and F1 scores of SikuGPT on each category are shown in Table 9. Among the 14 categories, SikuGPT obtained high scores in categories such as astronomy, medicine, and agriculture, while performing relatively poorly in categories such as miscellaneous literature, Taoism, and Legalism. This may be related to the characteristics of the texts in each category. The texts in the astronomy category often exhibit salient features in describing astronomical phenomena, such as "平帝元始元年辛酉歲五月丁巳朔日食"(This sentence means: "The solar eclipse occurred in the early morning of Dingsi day in May of Xinyou year in the first year of Yuanshi of Emperor Ping."). By contrast, the texts in the miscellaneous literature and Taoism categories encompass a wide range of content and structure, without exhibiting obvious features in text content and structure. Additionally, the poor performance of the Legalism category may be due to the small size of the training set, which affected the model's performance on the classification task.

**Table 9 Classification effect of each category of the best performing model SikuGPT**

| Category | Precision | Recall | F1-score |
| --- | --- | --- | --- |
| 儒家（Confucianism） | 82.85 | 89.18 | 85.90 |
| 兵家（Military Strategists） | 93.42 | 88.91 | 91.11 |
| 农家（Agriculturalism） | 93.43 | 94.83 | 94.12 |
| 医家（Medical Science） | 92.97 | 97.82 | 95.33 |
| 天文算法（Astronomy and Mathematics） | 98.30 | 96.56 | 97.43 |
| 小说家（Fiction） | 88.53 | 85.21 | 86.84 |
| 术数（Divination and Numerology） | 89.24 | 94.90 | 91.98 |
| 杂家（Eclecticism） | 79.12 | 71.85 | 75.31 |
| 法家（Legalism） | 82.44 | 87.10 | 84.71 |
| 类书（Genus-books） | 94.30 | 91.32 | 92.79 |
| 艺术（Arts and Crafts） | 94.05 | 93.57 | 93.81 |
| 谱录（Rituals and Records） | 91.90 | 91.81 | 91.85 |
| 道家（Taoism） | 84.71 | 86.77 | 85.73 |
| 释家（Buddhism） | 94.00 | 93.08 | 93.54 |

# 5 Discussion

The validation experiments in this study demonstrate the relative superiority of SikuGPT in downstream tasks such as text translation and text classification, providing the following insights:

## 5.1 Model domain specialization: an effective strategy for vertical domain tasks

In this study, pre-training GPT2-chinese-ancient was continued on the ancient Chinese corpus of the *Siku Quanshu* to obtain SikuGPT, which showed significantly improved performance in natural language processing tasks such as ancient Chinese text translation and text classification compared to the baseline model. This result indicates that using domain-specific corpora to continue pre-training generative models can effectively improve their performance in executing downstream tasks on texts with similar linguistic and domain characteristics. Furthermore, in this study the original vocabulary of GPT2-chinese-ancient was enlarged with over 5000 traditional Chinese characters. The purpose was to enhance the model's encoding ability for ancient Chinese texts and avoid the model's inability to recognize rare characters and generate invalid characters such as "[UNK]". Currently, improvement of models' performance heavily relies on the quality and scale of training corpora. The superior performance of SikuGPT in this study further demonstrates that building domain-specific models represents an effective and necessary means of improving model performance on vertical domain tasks. On the basis of the huge success achieved by universal generative language models, large models specifically designed for a given domain will show even greater performance advantages. However, the construction of high-quality large-scale corpora at lower costs still poses an essential research problem in current information resource management disciplines.

## 5.2 Generative pre-training models: a new paradigm for intelligent processing of classics

In this study, a generative model was applied to text translation, which promoted the innovation of the research paradigm for text translation. Traditional translation models consist of two core modules: an encoder that encodes the source language text to a semantic vector and a decoder that receives the vector and decodes it into natural language text expressed in the target language. Models based on the encoder-decoder architecture require a large amount of parallel corpora consisting of source and target language texts during the training phase. Consequently, the effect of the translation model depends heavily on the granularity and scale of the aligned parallel corpus. In contrast, the various GPT models used in this study serve as decoders with a simpler structure than those of traditional translation models. Moreover, the construction of the translation model in this study was realized through two steps of continued pre-training and fine-tuning. During pre-training, only the traditional Chinese corpus of *Siku Quanshu* was utilized, avoiding the cumbersome task of constructing a large number of parallel corpora while improving the model's performance on translation tasks. It is foreseeable that if bilingual or multilingual corpora are exploited during pre-training, the model's performance on downstream tasks such as translation can be further improved.

This study also demonstrates the feasibility of applying generative models to text classification tasks. Text classification models have evolved from traditional statistical machine learning models through traditional deep learning models to pre-trained language models such as BERT. The mainstream approach in the past has been to utilize Encoder-based pre-trained models such as BERT combined with an additional output layer to build text classification models. However, models constructed through this approach have limited generality, as they can only be used for text classification and can only classify text into a limited number of categories. In this study, a Decoder-based generative model was adopted to complete the text classification task. The generative model uses the text to be classified as a "prompt" for the model, which then generates content that can serve as a category label. The model achieved accuracy comparable to that of vertical domain models, which demonstrates that generative models can reasonably generate expected content under the joint constraint of the task corpus and prompt template. Additionally, for the original pre-trained language model, this model can effectively avoid its semantic representation ability being weakened during the fine-tuning stage in the traditional "pre-training + fine-tuning" mode.

This study also demonstrates that generative models can be employed to complete text classification tasks without the need for additional output layers. This stands in contrast to the mainstream approach using pre-trained Encoder-based models such as BERT. Generative models can be applied to multiple tasks, including text translation and text classification, and have a wide range of applicability. This capability may reshape the paradigm of organizing and serving knowledge of ancient texts.

## 6. Conclusion

In this study, the SikuGPT pre-training model was developed by incorporating the corpus of *Siku Quanshu* into the GPT model. The SikuGPT pre-training model is a practical application of the integration and development of generative pre-training technology and digital humanities research. It not only expands the application field of generative pre-training technology, but also enriches the technical content of digital humanities. In light of this, the study has major theoretical and practical significance for ancient text processing and digital humanities research.

In the future, we will closely follow the latest development in generative AI research, and explore the training of pre-training models with larger parameter scales and efficient parameter fine-tuning. In addition, new technologies and concepts need to be actively integrated in the field of natural language processing to deeply explore the value and knowledge of traditional Chinese culture. Furthermore, a knowledge service platform can be developed based on large pre-training models for ancient Chinese. With such platform, multiple NLP tasks of ancient texts can be unified through a question-answering system, thus providing users with high-quality ancient knowledge services. Ultimately, this will promote the dissemination of traditional Chinese culture and the knowledge contained in ancient Chinese texts. Finally, we open source the sikuGPT model to github (https://github.com/SIKU-BERT/sikuGPT) so that researchers from different professional backgrounds can download it and jointly promote the communication between different cultures.

## Acknowledgements

The authors acknowledge the National Social Science Foundation of China (Grant Numbers: (21&ZD331)) for financial support. We thank all the volunteers and all publications support and staff who wrote and provided helpful comments on previous versions of this document.

## Conflict of interest statement

The authors declared that they have no conflicts of interest to this work.